\documentclass[conference]{IEEEtran}
\IEEEoverridecommandlockouts
\usepackage{url}
\usepackage{cite}
\usepackage{amsmath,amssymb,amsfonts}
\usepackage{algorithmic}
\usepackage{graphicx}
\usepackage{textcomp}
\usepackage{xcolor}
\usepackage[caption=false,font=footnotesize]{subfig}  
\usepackage{hyperref}
\usepackage{cleveref}
\usepackage{amssymb}

\def\BibTeX{{\rm B\kern-.05em{\sc i\kern-.025em b}\kern-.08em
    T\kern-.1667em\lower.7ex\hbox{E}\kern-.125emX}}
\begin{document}

\title{UruDendro4: A Benchmark Dataset for Automatic Tree-Ring Detection in Cross-Section Images of \textit{Pinus taeda L.}\\
\thanks{This work was funded by ANII under project number ANII-FMV-176061.\\ 979-8-3315-9170-0/25/\$31.00 ©2025 IEEE}}

\author{
\IEEEauthorblockN{Henry Marichal}
\IEEEauthorblockA{\textit{Ingeniería Eléctrica} \\
\textit{Universidad de la República}\\
Montevideo, Uruguay \\
hmarichal93@gmail.com}
\and
\IEEEauthorblockN{Joaquín Blanco}
\IEEEauthorblockA{\textit{Ingeniería Forestal} \\
\textit{Universidad de la República}\\
Tacuarembó, Uruguay \\
joaquinblanco77@gmail.com}
\and
\IEEEauthorblockN{Diego Passarella}
\IEEEauthorblockA{\textit{Proc. Ind. de la Madera} \\
\textit{Universidad de la República}\\
Tacuarembó, Uruguay \\
diego.passarella@cut.edu.uy}
\linebreakand 
\IEEEauthorblockN{Gregory Randall}
\IEEEauthorblockA{\textit{Ingeniería Eléctrica} \\
\textit{Universidad de la República}\\
Montevideo, Uruguay \\
randall@fing.edu.uy}
}



\maketitle

\begin{abstract}
Tree-ring growth represents the annual wood increment for a tree, and quantifying it allows researchers to assess which silvicultural practices are best suited for each species. Manual measurement of this growth is time-consuming and often imprecise, as it is typically performed along 4 to 8 radial directions on a cross-sectional disc. In recent years, automated algorithms and datasets have emerged to enhance accuracy and automate the delineation of annual rings in cross-sectional images.

To address the scarcity of wood cross-section data, we introduce the \textit{UruDendro4} dataset—a collection of 102 image samples of \textit{Pinus taeda L.}, each manually annotated with annual growth rings. Unlike existing public datasets, \textit{UruDendro4} includes samples extracted at multiple heights along the stem, allowing for the volumetric modeling of annual growth using manually delineated rings. This dataset (images and annotations) allows the development of volumetric models for annual wood estimation based on cross-sectional imagery. 

Additionally, we provide a performance baseline for automatic ring detection on this dataset using state-of-the-art methods. The highest performance was achieved by the DeepCS-TRD method, with a mean Average Precision of 0.838, a mean Average Recall of 0.782, and an Adapted Rand Error score of 0.084. A series of ablation experiments were conducted to empirically validate the final parameter configuration. Furthermore, we empirically demonstrate that training a learning model including this dataset improves the model's generalization in the tree-ring detection task.
\end{abstract}


\section{Introduction}
\begin{figure}
    \centering
    \includegraphics[width=\linewidth]{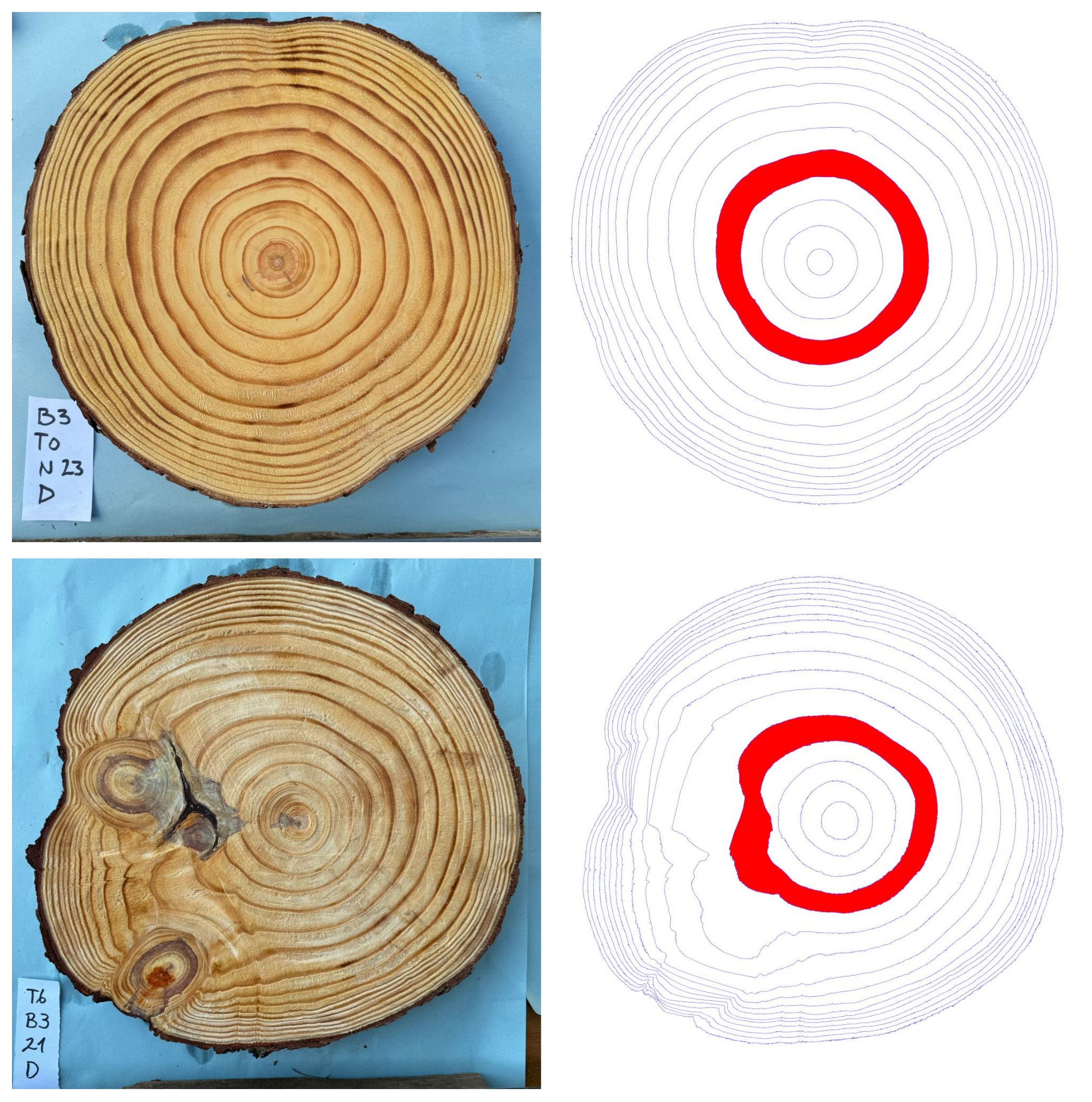}
    \caption{Cross-section samples of  \textit{Pinus taeda L.}  (\texttt{T0\_B3\_N23\_D} and \texttt{T6\_B3\_N21\_D} from \textit{UruDendro4}, respectively) and their tree-ring curve delineations. The fifth growth ring area is highlighted in red in both samples.}
    \label{fig:main}
\end{figure}

The annual growth of certain tree and shrub species produces a distinct pattern in the wood known as annual growth rings. In a transverse cross-section of the trunk, these rings appear as concentric shapes centered on the innermost ring, referred to as the pith. The outermost ring corresponds to the last year of growth. \Cref{fig:main} illustrates two samples of \textit{Pinus taeda L.} cross-sections and their delineated tree-ring curves.

In commercial forestry, where the primary objective is to maximize timber production, silvicultural practices are applied and evaluated to improve growth. These practices must be studied over long periods; for example, the samples in \Cref{fig:main} were subjected to different plantation densities and thinning procedures. By analyzing how specific silvicultural treatments influence tree growth year by year, it is possible to identify which practices are effective or inadequate under a given climatic and soil conditions and expected wood quality.

A common approach to assessing this relationship involves measuring tree-ring widths on transverse stem cross-sections (discs), typically along four to eight fixed radial directions starting from the pith position. These measurements are often taken manually or semi-automatically from images, such as those shown in \Cref{fig:main}. However, this one-dimensional approach may provide limited insights. Several studies have demonstrated stronger correlations between ring-based metrics and climate variables, including temperature and precipitation. For instance, Labrecque-Foy et al.~\cite{basal_area} found that computing the basal area increment (BAI) from ring-width data in shrubs yielded a better correlation with climate signals than using raw ring-width series.

Extracting two-dimensional features from tree-ring patterns, such as the area enclosed between successive rings (highlighted in red in \Cref{fig:main}), may offer more accurate information on silvicultural effects and wood production. However, obtaining valuable 2D measurements from disc images requires precise delineation of annual ring boundaries. Manually performing this task is extremely time-consuming. Although several open-source algorithms have been developed for automatic ring detection, a manual postprocessing step is still required to add missing rings or correct false detections.

To increase the diversity of publicly available data and support the development of more robust tree-ring detection algorithms in cross-section images, we make the following contributions:
\begin{itemize}
    \item We present \textbf{UruDendro4}, a publicly available dataset containing 102 annotated images of \textit{P. taeda} cross-sections, supporting ring boundary detection.
    \item We describe the image acquisition protocol, annotation methodology, and characteristics of the dataset.
    \item We benchmark recent ring boundary detection algorithms on the dataset to provide a baseline and highlight the challenges posed by the data.
\end{itemize}

\section{Related work}
The first publicly available dataset of 64 images of \textit{P. taeda} with manual annual ring delineation, named \textit{UruDendro}, was introduced in \cite{uru1}.
An automatic method for tree ring delineation in cross-section images, named Cross-Section Tree-Ring detector (CS-TRD),  was proposed by the same authors in \cite{cstrd}. The CS-TRD method operates by detecting edges corresponding to annual tree ring boundaries and carefully recombining them using classical image processing techniques. A second dataset of 53 images of \textit{P. taeda} and manual annual ring delineation, named \textit{UruDendro2}, was introduced in \cite{deepcstrd} together with a deep-learning modification of the CS-TRD method, which improved the accuracy of the original method in \textit{P. taeda}. This modification, named DeepCS-TRD, replaces the edge detection step in CS-TRD with the U-Net deep learning architecture \cite{unet}. Additionally, they released the \textit{UruDendro3a} dataset, which consists of nine annotated cross-section samples of the \textit{Gleditsia triacanthos} species.


Kennel et al. \cite{KennelBS15} included a dataset of seven cross-section images of \textit{Abies alba}, although the ring annotations are not fully accessible. Additionally, the authors proposed an algorithm based on solving a partial differential equation, incorporating terms related to both the image content and the curve itself, to automatically delineate the tree rings. However, the code is not available.

Gillert et al. \cite{inbd} introduced three datasets of microscopy images of shrubs of the \textit{Dryas octopetala}, \textit{Empetrum hermaphroditum}, and \textit{Vaccinium myrtillus} species, with 213 delineated samples.  The Iterative Next Boundary Detection method (INBD) was presented in the same work. It is a deep-learning approach developed for tree ring detection in microscopy images of shrubs, demonstrating superior performance compared to the CS-TRD method in the species \textit{P. taeda} when trained properly \cite{deepcstrd}. This method comprises two steps: first, the background, ring boundaries, and center region are segmented. Secondly, a refinement stage is used, where patches are extracted iteratively from the inner to the outer rings, segmenting each ring individually. Both steps utilized a U-Net architecture, which required separate training.  Another microscopy dataset of the \textit{Salix glauca} shrub species was released in \cite{disko}, comprising 50 image samples with their corresponding annual ring annotations. 

Regarding image datasets with no manual ring annotations, Longuetaud et al. \cite{DouglasFir} released the $TreeTrace\_Douglas$ database in November 2022, including several wood cross-section image collections. This dataset includes images and measurements, such as growth ring width and pith pixel location, along with other wood properties, which were acquired at various stages of the 
processing in a sawmill. 
A few months later, in February 2023, Longuetaud et al. \cite{PiceaAbis} introduced the $TreeTrace\_Spruce$ database, which contains images and measurements of 100 Norway spruce samples. While both databases are valuable contributions to the forestry community, they lack digital tree-ring annotations for each image, which are needed to assess computer vision algorithms.

\Cref{tab:dataset_all} summarizes the public annotated cross-section datasets. 
\textit{UruDendro4}, introduced here, is the largest in terms of the number of samples and the number of annotated rings.  In addition, the dataset contains systematic information of different thinning schemes, allowing for the assessment of the effect of different silvicultural practices on the growth of \textit{P. taeda}.


\begin{table}
\centering
\caption{Image Cross-Section datasets annotations summary. Number of samples (NB), number of rings (NR), tree or shrub species, and image acquisition method.}
\label{tab:dataset_all}
\resizebox{\linewidth}{!}{
\begin{tabular}{l|c|c|c|c}
\textbf{Dataset} & \textbf{NB} & \textbf{NR} & \textbf{Species} & \textbf{Acquisition} \\ \hline
Kennel et al. \cite{KennelBS15}                        & 7   & 212  & \textit{Abies alba}           & Scanner              \\
UruDendro3a \cite{deepcstrd}                            & 9   & 219  & \textit{G. triacanthos}       & Professional Camera  \\
Gillert et al. \cite{inbd} VM                           & 67  & 494  & \textit{V. myrtillus}         & Scanner              \\
Gillert et al. \cite{inbd} DO                           & 66  & 544  & \textit{D. octopetala}        & Scanner              \\
C. C. Power et al. \cite{disko}                        & 50  & 654  & \textit{S. glauca}            & Scanner              \\
Gillert et al. \cite{inbd} EH                           & 82  & 949  & \textit{E. hermaphroditum}    & Scanner              \\
UruDendro2 \cite{deepcstrd}                            & 53  & 1151 & \textit{P. taeda}             & Smartphone           \\
UruDendro \cite{uru1}                                  & 64  & 1630 & \textit{P. taeda}             & Smartphone           \\
\textbf{UruDendro4}                                    & 102 & 1930 & \textit{P. taeda}             & Smartphone           
\end{tabular}
}
\end{table}

\section{The \textit{UruDendro4} Dataset}

In this work, we introduce the \textit{UruDendro4} dataset, a new collection of 102 high-resolution images with expertly annotated tree-ring boundaries. Compared to previous collections, 
this new release nearly doubles the number of annotated \textit{P. taeda} samples, increasing from 117 (64 samples form  \textit{UruDendro} \cite{uru1} and 53 from \textit{UruDendro2} \cite{deepcstrd}) to 219.

A distinctive feature of \textit{UruDendro4} is that it includes multiple cross-sections from the same tree. This enables the estimation of annual volume increment per tree, expanding the possibilities for growth analysis beyond traditional one-slice ring-width measurements.

\begin{table*}
\centering
\caption{Description of silvicultural treatments: thinning schedule, tree densities before and after thinning, pruning height, and current stand density. The pre-thinning density for all treatments was 600 trees ha$^{-1}$.}
\label{tab:treatments_en}
\resizebox{\textwidth}{!}{%
\begin{tabular}{l|c|c|c|c|c}
\textbf{Treatment} & \textbf{Thinning timing [years]} &  \textbf{Post-thinning density [trees.ha$^{-1}$]} & \textbf{Pruning height [m]} & \textbf{Current density [trees.ha$^{-1}$]} \\ \hline
T0 – Commercial management & 11 &  350 & between 5.5 and 8.0 & 350 \\ 
T2 – Heavy thinning / High pruning & 8 &  250 & 8.0 & 250 \\ 
T4 – Current thinning / High pruning & 8  & 350 & 8.0 & 350 \\ 
T6 – Two thinnings / High pruning & \begin{tabular}[c]{@{}c@{}}1st thinning: 8\\ 2nd thinning: 18\end{tabular}  & \begin{tabular}[c]{@{}c@{}}1st thinning: 400\\ 2nd thinning: 200\end{tabular} & 8.0 & 200 \\
\end{tabular}
}
\end{table*}


The trees were collected as part of a study examining the impact of various silvicultural practices on wood production.  The treatments are summarized in \Cref{tab:treatments_en}. The experiment was a randomized complete block design, with block identifiers labeled as \texttt{B1} to \texttt{B3}.  The discs were collected between April and July of 2024.

\subsection*{Harvest Site}

102 cross-sections of \textit{P. taeda} were collected from the site known as "La Altura", located approximately 4 km from the city of Tranqueras (31°13'30.2'' S, 55°46'53.6'' W), in the Department of Rivera, Uruguay. Trees were felled using a chainsaw, and four discs per stem were extracted at heights of 0m, 3m, 6m, and 9m. An additional disc was collected at 1.3m for the T0 treatment.

The soils in the area are very deep, with limited drainage, a loamy-sandy texture, and extremely low fertility (a productivity index of 53). The dominant soils are Umbric Albic Luvisols, classified as priority forestry soils, belonging to the 7.42 group according to CONEAT scale (National Committee of Agronomic Studies of Soils in Uruguay). The terrain consists of gently rolling hills with slopes of up to 3\%.

The meteorological data from the site were provided by the Uruguayan Institute of Meteorology (INUMET). They include minimum, mean, and maximum temperature records [\textdegree C] and cumulative precipitation [mm] from January 2002 to April 2024. Temperature data were recorded at the Rivera Meteorological Station (30\textdegree53'47.3''S, 55\textdegree32'33.6''W), the closest station to the test site. Mean temperatures represent the daily measurements at hourly intervals. Precipitation data were obtained about 4 km from the study area. Data are presented as daily, monthly, and annual averages (for temperatures, including extremes) and as daily, monthly, and annual cumulative values (for precipitation).

\subsection*{Image Acquisition}

\begin{figure*}
    \centering
    \subfloat[T0\_B1\_N27\_C]{%
        \includegraphics[width=0.3\textwidth]{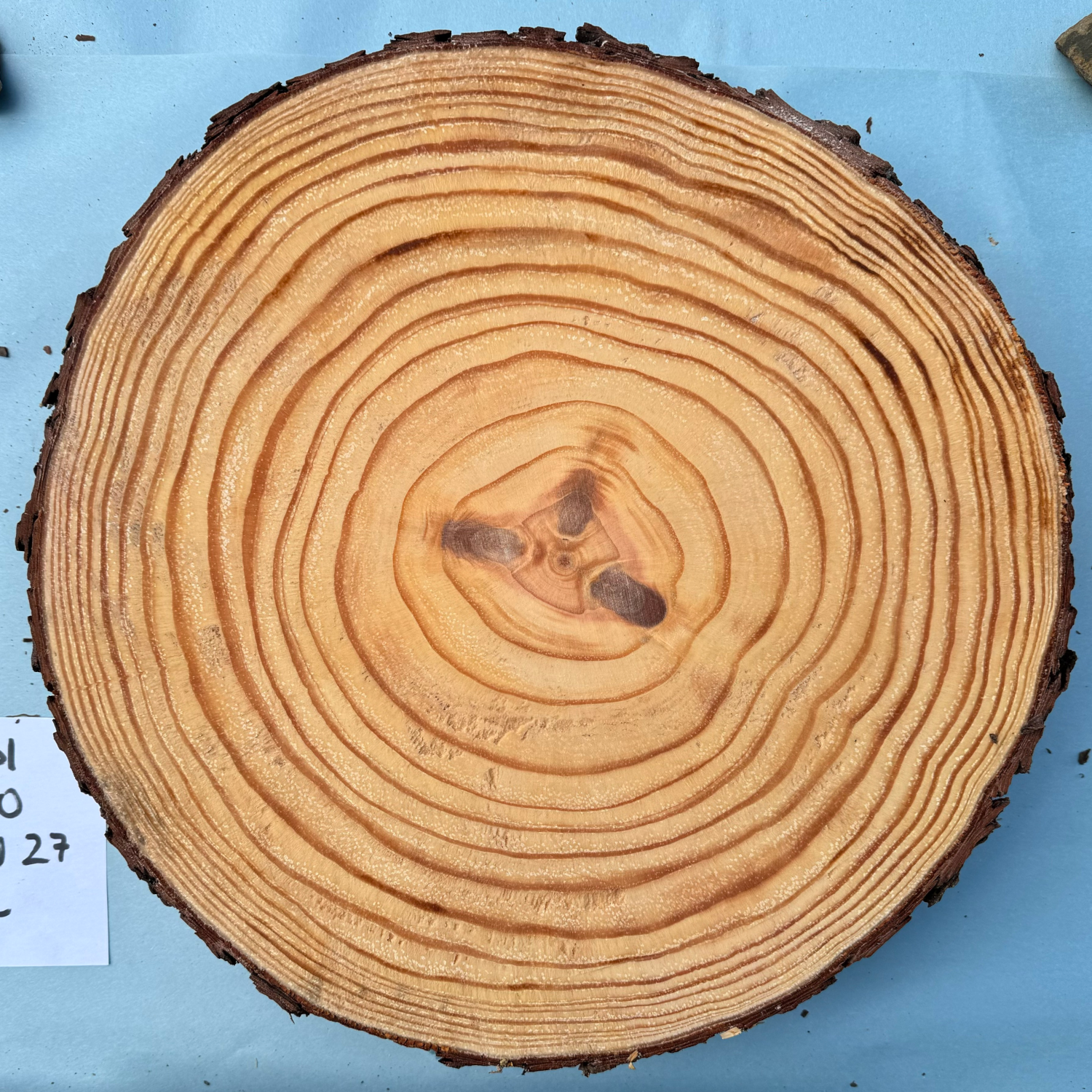}%
        \label{fig:subfig_a}}
    \hfill
    \subfloat[T2\_B3\_N9\_C]{%
        \includegraphics[width=0.3\textwidth]{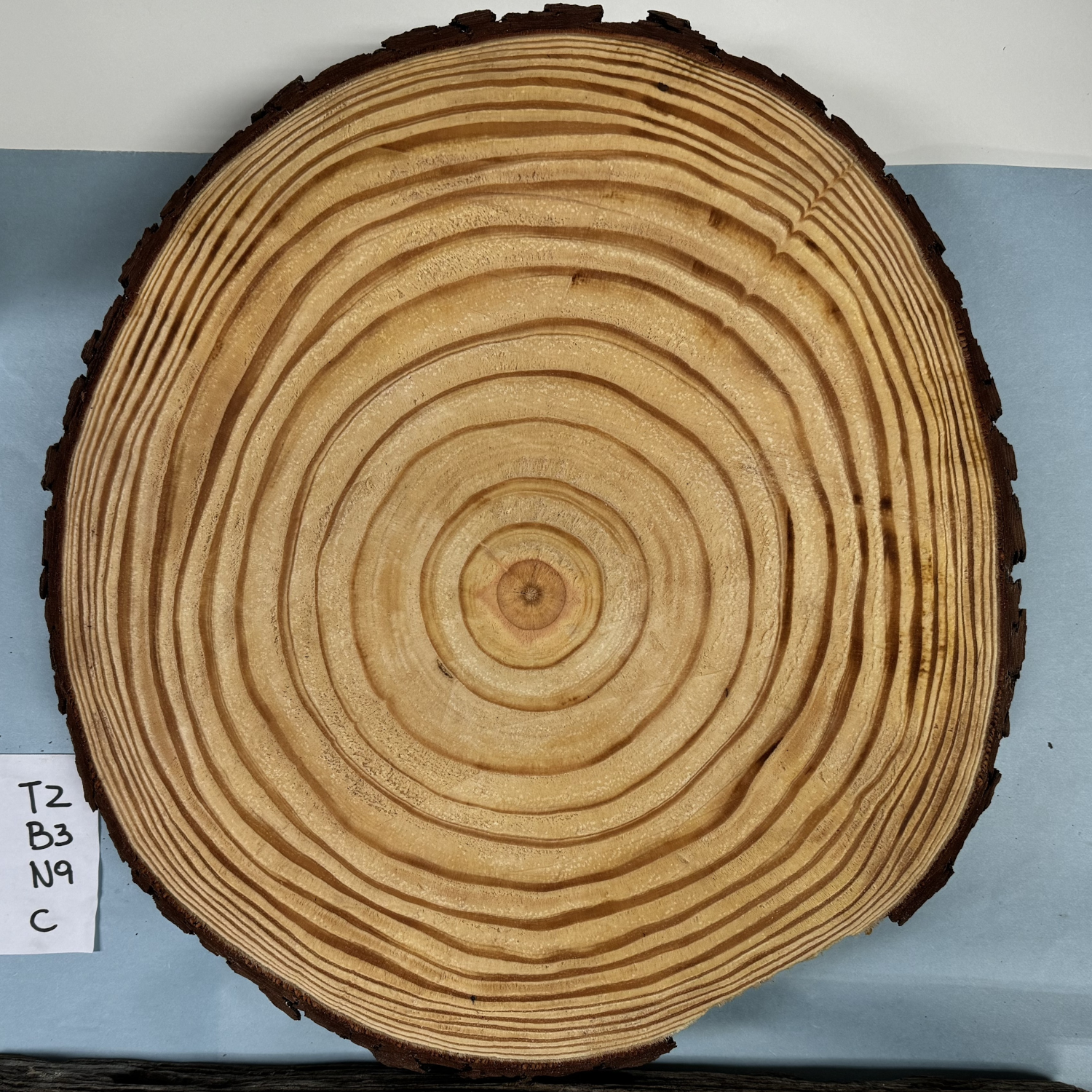}%
        \label{fig:subfig_b}}
    \hfill
    \subfloat[T0\_B1\_N32\_ADAP]{%
        \includegraphics[width=0.3\textwidth]{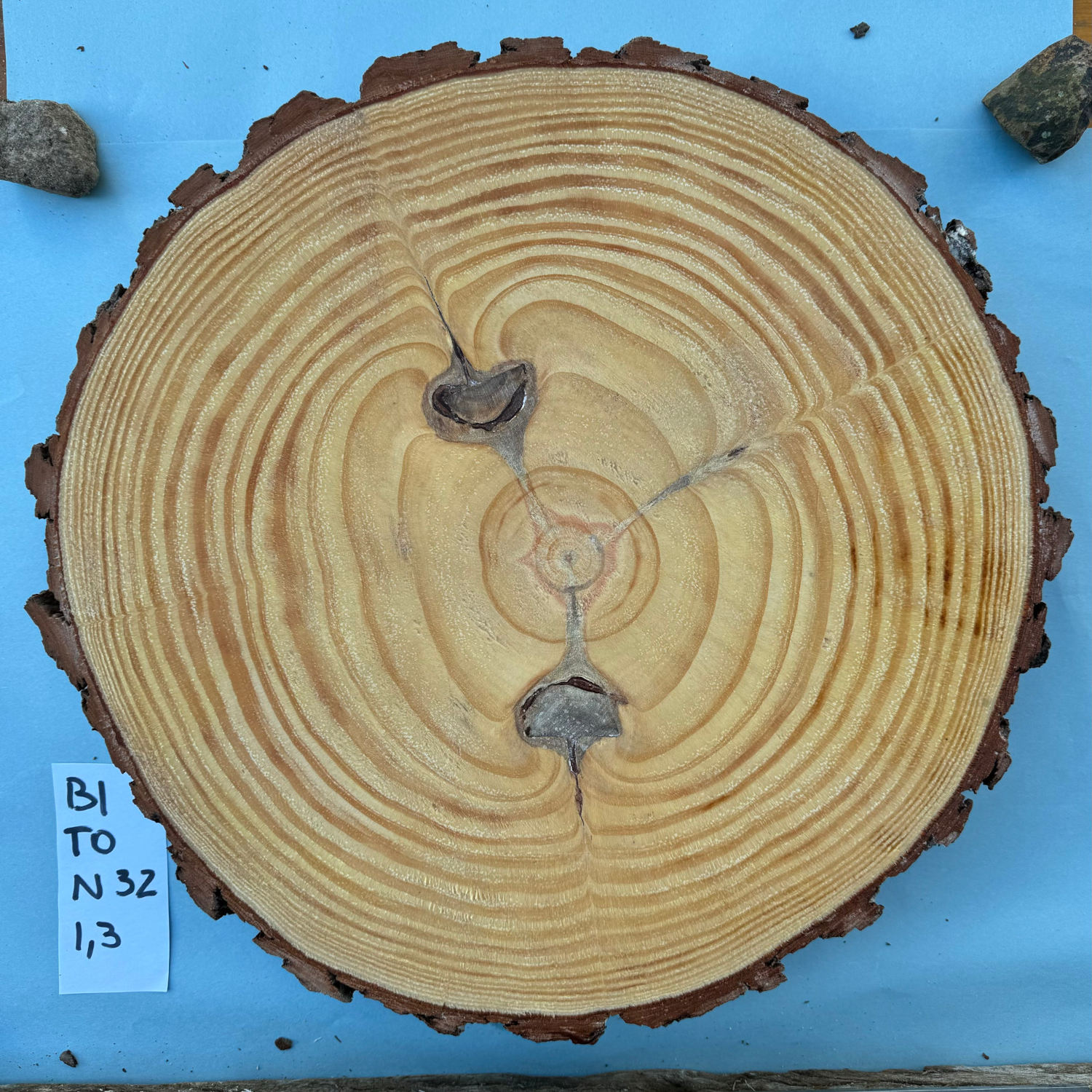}%
        \label{fig:subfig_c}}
    \caption{Example images from the \textit{UruDendro4} dataset.}
    \label{fig:urudendro4}
\end{figure*}

The discs were polished using an electric planer. 
The discs were moistened before image capture. Photographs were taken under varying lighting conditions: samples T2 and T4 (B1 and B2) were photographed indoors under artificial white light; T6 (B2) samples were photographed outdoors; T4 (B3), T6 (B1 and B3), and T0 samples were photographed outdoors in shaded conditions. All photographs were taken using an iPhone 15 Pro smartphone. The minimum image width is 1317 pixels, and the maximum width is 4695 pixels. \Cref{fig:urudendro4} illustrates some examples from the dataset.

Sample names follow the structure:
\[
TX\_BY\_NZ\_W
\]
Where $TX$ indicates the treatment applied to the tree (0, 2, 4, or 6); $BY$ refers to the block (1, 2, or 3); $NZ$ refers to the tree identifier (e.g., 3, 12, 23, etc.); and $W$ refers to the sample height (A = between 10\ and 
30 cm, B = 3m, C = 6m, D = 9m, and ADAP = 1.3m). 

\subsection*{Tree-Ring Annotation Procedure}

Annual ring annotations were made using the Labelme tool \cite{labelme}. Each annual ring is represented as a continuous curve formed by connected pixels. Rings are roughly concentric around the pith, i.e., they don't intersect with each other, and each annual ring fully encloses the rings from previous years.

Initial annotations were automatically generated using the CS-TRD method \cite{cstrd}, in combination with the APD method for pith detection \cite{apd}. These initial predictions were subsequently post-processed and corrected by a human annotator to ensure the accuracy of the ring delineations.

\subsection*{File structure}

The dataset \cite{dataset} is available at the Zenodo Repository at \url{https://doi.org/10.5281/zenodo.15653340}. It is organized into several files containing images and marks as described below:

\begin{itemize}
    \item images/ - Folder with the image samples
    \item images\_no\_background/ - Folder with the images after the background has been removed
    \item annotations/
    \begin{itemize}
        \item annual\_rings/ - Annual ring annotations
    \end{itemize}
    \item pith\_location.txt - file with the pith pixel location for each image
\end{itemize}

The name of each sample serves as its code. For example, sample T4\_B3\_N1\_C  has three different files: \
\begin{itemize}
    \item images/T4\_B3\_N1\_C.png
    \item images\_no\_background/T4\_B3\_N1\_C.jpg
    \item annotations/annual\_rings/T4\_B3\_N1\_C.json
\end{itemize}

\section{Methodology}
We conduct a series of ablation experiments using our dataset to evaluate the automatic detection of tree rings in cross-section images. Details on the experimental setup and the datasets used are provided in the following sections.

\subsection*{Baselines}

We evaluate the performance of 3 state-of-the-art methods for the automatic detection of annual ring curves.

\textit{CS-TRD} \cite{cstrd}: A classical image processing approach for ring-edge detection. The parameters were set to $\sigma = 3.0$ and $\alpha = 45$. No internal resizing was applied to the input images.

\textit{DeepCS-TRD} \cite{deepcstrd}: This deep learning-based model was trained using a batch size of 8 for 100 epochs. The Adam optimizer was employed with an initial learning rate of $1 \times 10^{-3}$, and a cosine annealing schedule was used to dynamically adjust the learning rate throughout training. The Dice loss function was adopted to optimize the segmentation performance. A ResNet18 backbone pre-trained on ImageNet was used to enhance feature extraction in the encoder. The model weights corresponding to the lowest validation loss were selected as the final model. No tile overlap strategy was applied during training (i.e., no patching, with the tile\_size parameter set to 0). All parameters, including the encoder weights, were optimized during training unless explicitly stated otherwise.

\textit{INBD} \cite{inbd}: Both stages of the INBD model were trained under the same configuration. A downsampling factor of 2 was applied, meaning that the input images were internally resized to half their original resolution. Each network was trained for 100 epochs using a batch size of 8, the AdamW optimizer, an initial learning rate of $1 \times 10^{-3}$, and a cosine annealing learning rate schedule. Only the decoder and classifier layers were trainable. A MobileNetV3 backbone, pre-trained on ImageNet, was used to enhance feature extraction in the encoder.

All models were trained on a NVIDIA A40 GPU with 48 GB of VRAM, utilizing the infrastructure provided by ClusterUY \cite{clusterUy} (site: https://cluster.uy).

\subsection*{Data}

The \textit{UruDendro4} dataset is divided into training, validation, and test sets using a 60\%/20\%/20\% split, resulting in 62, 20, and 20 images, respectively. All performance results reported in this work are based on the test subset.

Additionally, for cross-domain validation, the DeepCS-TRD method was trained on the \textit{UruDendro} \cite{uru1} and \textit{UruDendro2} \cite{deepcstrd} datasets, which contain 64 and 53 annotated images, respectively. All images from each dataset were used for training in the cross-domain experiments.

Moreover, we annotated a small subset of two images for the \textit{Douglas fir} species \cite{DouglasFir} (79 tree rings), aiming to assess the model's generalization capabilities to new species.

\textit{Preprocessing:} The same preprocessing protocol described in \cite{deepcstrd} was applied to the images. First, the background was removed using the $U^2$-Net network \cite{u2net}. 
In addition, the background margins were cropped to focus on the disc area, ensuring a minimum distance of 50 pixels between the wood cross-section and the image borders. Finally, images were resized to $1504$ pixels on the highest dimension using Lanczos interpolation.


\section{Experimental results}


\subsection*{Metrics}
Results are reported on the \textit{UruDendro4} test set using the same input resolutions as during training. We report the mean Average Precision ($m$AP) and the mean Average Recall ($m$AR) metrics. These metrics are averaged over multiple Intersection-Over-Union (IoU) thresholds, from 0.5 to 1.0 with a step of 0.05. Additionally, we reported the Adapted Rand error (ARAND) using the \texttt{adapted\_rand\_error} method from \texttt{skimage} (version 0.25). Metrics are reported as the mean and standard deviation over five independent experiments.

\subsection*{Quantitative Evaluation}

The three automatic tree ring detection methods, CS-TRD \cite{cstrd}, DeepCS-TRD \cite{deepcstrd}, and INBD \cite{inbd} are assessed over the \textit{UruDendro4} test set. In all cases, the pith position was automatically estimated with the Automatic Wood Pith Detector (APD) method \cite{apd}. 

\Cref{tab:baseline} illustrates the results. DeepCS-TRD and INBD are deep-learning methods based on the U-Net architecture, 
trained on the \textit{UruDendro4} training set. The CS-TRD method is a classic image processing algorithm that doesn't require training. DeepCS-TRD method trained in the \textit{UruDendro4} dataset achieved the best results in all the metrics: $0.829\pm0.006$ mAP, $0.769\pm0.004$ mAR, and $0.094\pm0.007$ ARAND.

\begin{table}
\centering
\caption{Evaluation of the methods on the \textit{UruDendro4} dataset. 
Encoder weights were frozen in both INBD and DeepCS-TRD.}
\label{tab:baseline}
\begin{tabular}{l|c|c|c}
\textbf{Method}      & \textbf{mAP} ($\uparrow$)  & \textbf{mAR} ($\uparrow$)  & \textbf{ARAND} ($\downarrow$) 
\\ \hline
CS-TRD      & .576 (.000) &.568 (.000) & .174 (.000) \\
INBD       & .736 (.006)  &.712 (.004)   &   .099 (.003)  \\
DeepCS-TRD  & \textbf{.831 (.007)} &\textbf{.775 (.007)} & \textbf{.087 (.004)} \\
\end{tabular}
\end{table}

The INBD method, despite being developed for ring detection in microscopy images of shrubs, achieved remarkable performance, albeit not as good as that of DeepCS-TRD. Compared to the DeepCS-TRD method, it required a longer training time (18 hours vs. 33 minutes).


Furthermore, we evaluated the performance of automatic tree ring detection by incrementing the dataset training size and employing different training strategies. We used the DeepCS-TRD method because it required less training time than the INBD and achieved better performance. \Cref{tab:deep_cstrd_assestment} illustrates the results. First, we evaluated the cross-domain capabilities by training the method with the \textit{UruDendro} (Uru) dataset, the \textit{UruDendro2} (Uru2) dataset, and the union of both datasets (Uru + Uru2). In all the cases, the DeepCS-TRD models performed better than the CS-TRD method (see \Cref{tab:baseline}). The best results were obtained when the model was trained using both the \textit{UruDendro1} and \textit{UruDendro2} datasets, indicating that increasing the dataset size and diversity enhances the model’s ability to generalize to new domains.

\begin{table}
\centering
\caption{Evaluation of the DeepCS-TRD method over the test set in \textit{UruDendro4}, increasing the training size. The column's Dataset indicates the dataset used for training the method. Uru stands for \textit{UruDendro}, Uru2 for \textit{UruDendro2}, and Uru4 refers to the training subset of the \textit{UruDendro4} dataset. Additionally, we made experiments freezing the encoder layers during training (\checkmark). 
}
\label{tab:deep_cstrd_assestment}
\begin{tabular}{l|c|c|c|c}
\textbf{Dataset}     & \textbf{Freeze} &\textbf{mAP} ($\uparrow$)   & \textbf{mAR} ($\uparrow$)  & \textbf{ARAND} ($\downarrow$) \\ \hline 
Uru & &.710 (.010) & .683 (.011) &  .115 (.005) \\
Uru2 & &.694 (.009) &.680 (.005) & .112 (.005) \\
Uru + Uru2 & &.729 (.018)  &.705 (.015) & .107 (.007) \\\hline
Uru4 &  &.829 (.006) & .769 (.004) & .094 (.007) \\
Uru4 & \checkmark & \textbf{.831 (.007)} & \textbf{.775 (.007)} & .087 (.004) \\\hline
Uru + Uru2 + Uru4  & &.820 (.007) & .767 (.006) & \textbf{.087 (.003)} \\ 
Uru + Uru2 + Uru4  & \checkmark & .806 (.003) & .750 (.009) & .102 (.009)  \\ 
Uru + Uru2 \hspace{0.1cm}\vline \hspace{0.1cm}Uru4 & &.826 (.012) & .775 (.018) & .090 (.010)  \\ 
Uru + Uru2 \hspace{0.1cm}\vline \hspace{0.1cm}Uru4 & \checkmark& .826 (.010) & .774 (.007) & .087 (.006) \\ 

\end{tabular}
\end{table}

Next, we trained the model using the \textit{UruDendro4} training set, employing encoder weights pre-trained on ImageNet1k. Two experiments were conducted: in the first, all layers of the U-Net architecture (encoder, decoder, and classification head) were trained jointly; in the second, the encoder weights were frozen, and only the decoder and classification layers were updated during training. This allows for a reduction in the number of training parameters from  14.328.209 to 3.151.697. Freezing the encoder weights improved the results in all the metrics. 

We then tested different combinations of datasets (Uru, Uru2, and Uru4) and training strategies. In one approach, the model was trained using data from all three datasets simultaneously. In the alternative approach, we adopted a two-step training procedure: first, the model was trained for 100 epochs using the combined Uru and Uru2 datasets; then, starting from the weights that achieved the best performance on the Uru+Uru2 validation set, we fine-tuned the model for an additional 100 epochs using only the training subset of \textit{UruDendro4}. Among all the experiments presented in \Cref{tab:deep_cstrd_assestment}, the highest performance was achieved when training the model on the \textit{UruDendro4} dataset with frozen encoder weights. This suggests that the pre-trained encoder has strong feature extraction capabilities that generalize well to this task. 

The DeepCS-TRD method has a filtering edge parameter named $\alpha$, which is set to $45^{\circ}$ by default. We have tried different values in an attempt to improve the results. We have used the models trained with the \textit{UruDendro4} training set with frozen encoder weights (row 5 in \Cref{tab:deep_cstrd_assestment}). \Cref{tab:deep_cstrd_assestment_alpha} illustrates the results. Setting the $\alpha$ value to 60$^{\circ}$  increased the method's performance, achieving $0.838\pm0.016$ of mAP, $0.782\pm0.011$ of mAR, and $0.084\pm0.006$ of ARAND.

\begin{table}
\centering
\caption{DeepCS-TRD $\alpha$ parameter grid-search over the test set in \textit{UruDendro4}. Models trained on the Uru4 training set and using frozen encoder weights are employed (row 5 in \Cref{tab:deep_cstrd_assestment}).
}
\label{tab:deep_cstrd_assestment_alpha}
\begin{tabular}{l|c|c|c}
\textbf{$\alpha$($^{\circ}$)}      &\textbf{mAP} ($\uparrow$)   & \textbf{mAR} ($\uparrow$)  & \textbf{ARAND} ($\downarrow$) \\ \hline
30 & .799 (.014) & .734 (.022) & .121 (.017) \\ 
45 & .831 (.007) & .775 (.007) & .087 (.004)\\
60 & \textbf{.838 (.016)}& \textbf{.782 (.011)} & \textbf{.084 (.006)} \\
75 & .834 (.011) & .771 (.002) & .098 (.007) \\ 
90 & .789 (.006) & .694 (.695) & .125 (.121) \\ 
\end{tabular}
\end{table}

Finally, we aimed to assess whether the \textit{UruDendro4} dataset contributes to improved generalization of deep learning models in the tree-ring delineation task. To this end, we trained the DeepCS-TRD method using different combinations of the \textit{UruDendro1}, \textit{UruDendro2}, and \textit{UruDendro4} datasets. We evaluated the performance of the model using two samples from the \textit{Douglas fir} dataset \cite{DouglasFir}. The results, presented in \Cref{tab:generalization}, show that the model trained on a different species performs decently on another. The best performance was achieved when the model was trained using the combined dataset (Uru + Uru2 + Uru4).

\begin{table}
\centering
\caption{Evaluation of the DeepCS-TRD method over the \textit{Douglas fir} dataset when trained with \textit{P. taeda} images. 
}
\label{tab:generalization}
\begin{tabular}{l|c|c|c}
\textbf{Dataset}      &\textbf{mAP} ($\uparrow$)   & \textbf{mAR} ($\uparrow$)  & \textbf{ARAND} ($\downarrow$) \\ \hline
Uru & .629 (.054) & .623 (.060) & .208 (.041) \\
Uru2 & .725 (.025) & .688 (.037) & .176 (.039)  \\
Uru4 & .465 (.038) & .437 (.044) & .339 (.032) \\
Uru + Uru2 & .705 (.028) & .682 (.023) & .161 (.010) \\
Uru + Uru2 + Uru4 & \textbf{.728 (.013)} &\textbf{ .705  (.021)} & \textbf{.145 (.009)} \\ 
\end{tabular}
\end{table}

\subsection*{Qualitative Evaluation}

\begin{figure}
    \centering
    \includegraphics[width=\linewidth]{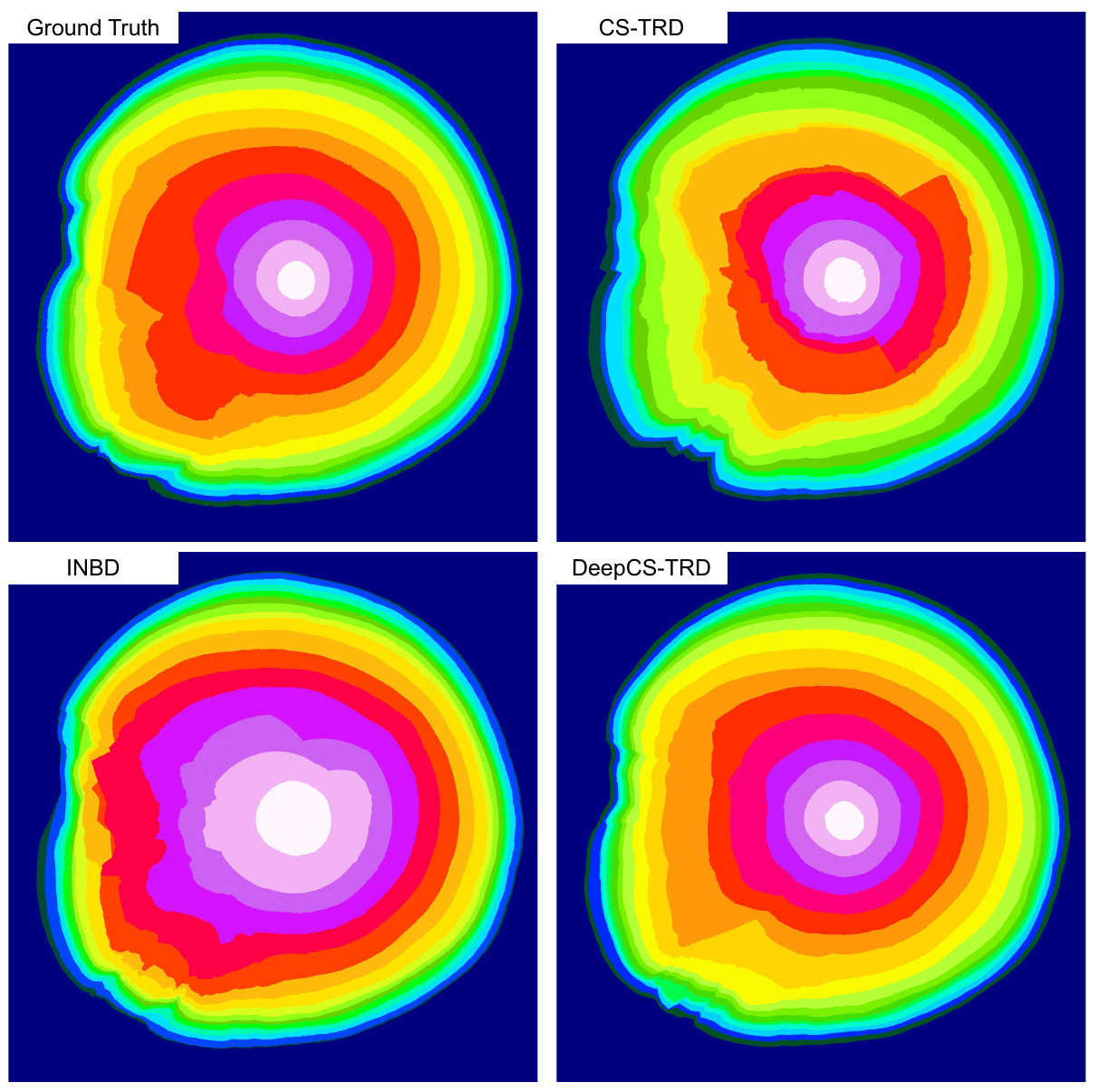}
    \caption{
    Automatic tree-ring detection results for sample \texttt{T6\_B3\_N21\_D} (see \Cref{fig:main}). 
    The CS-TRD method achieved a mAP of 0.256, a mAR of 0.241, and an ARAND score of 0.487. 
    INBD obtained 0.306 (mAP), 0.288 (mAR), and 0.302 (ARAND). 
    DeepCS-TRD outperformed both methods with scores of 0.688 (mAP), 0.688 (mAR), and 0.245 (ARAND), respectively.
    }
    \label{fig:t6_b3_n21_d}
\end{figure}

\begin{figure}
    \centering
    \includegraphics[width=\linewidth]{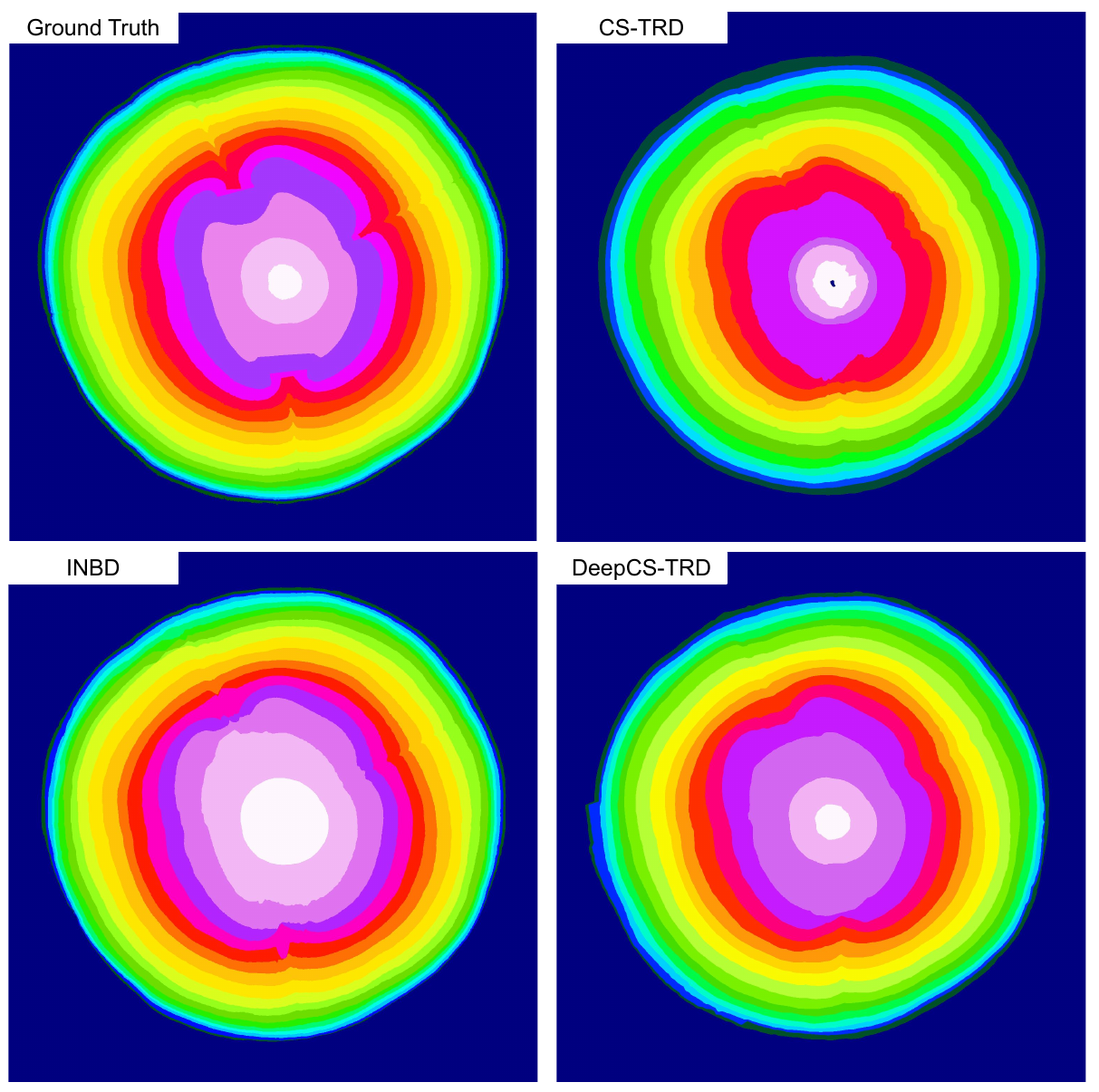}
    \caption{
    Automatic tree-ring detection results for sample \texttt{T0\_B1\_N32\_ADAP} (see \Cref{fig:urudendro4}). 
    The CS-TRD method achieved a mAP of 0.588, a mAR of 0.470, and an ARAND score of 0.200. INBD obtained 0.677 (mAP), 0.610 (mAR), and 0.135 (ARAND). 
    DeepCS-TRD outperformed both methods with scores of 0.800 (mAP), 0.680 (mAR), and 0.118 (ARAND), respectively.
    }
    \label{fig:to_b1_n32_adap}
\end{figure}

In this section, we qualitatively illustrate how the tree-ring automatic methods perform in the tree-ring detection task. \Cref{fig:t6_b3_n21_d} illustrates the tree-ring growth regions (Ground Truth subfigure) and the predicted tree-ring of each method for sample \texttt{T6\_B3\_N21\_D}. Despite the presence of two knots in the sample (see \Cref{fig:main}), the automatic methods performed notably well, with DeepCS-TRD achieving the best results. Results for sample \texttt{T0\_B1\_N32\_ADAP} (see \Cref{fig:urudendro4}) are presented at \Cref{fig:to_b1_n32_adap}. Again, the DeepCS-TRD method achieved more accurate tree-ring predictions.

\section{Conclusions}

Research aimed at improving both the quantity and quality of timber production in Uruguay is of great importance. Understanding how various silvicultural practices impact tree growth can help identify the most suitable practices for a specific soil type and given climatic conditions. 

Tree-ring delineation in cross-section images of \textit{P. taeda} requires high precision to enable meaningful correlations between ring-derived metrics and external factors such as temperature, precipitation, or silvicultural treatments. To support research in this area, we introduced the \textit{UruDendro4} dataset, which consists of 102 manually annotated cross-section images along with associated silvicultural and climate metadata, providing a valuable resource for forestry management and dendrochronology research.

We also demonstrated empirically that including this new dataset during training improves the generalization performance of state-of-the-art tree-ring delineation methods. Specifically, on the \textit{Douglas fir} dataset, the inclusion of \textit{UruDendro4} led to performance gains of 0.003 in mAP, 0.017 in mAR, and 0.031 in ARAND.

Finally, we conducted extensive ablation experiments using the DeepCS-TRD method, which established a strong baseline on \textit{UruDendro4} with scores of 0.838 in mAP, 0.782 in mAR, and 0.084 in ARAND.


\end{document}